\documentclass[conference]{IEEEtran}
\usepackage{fancyhdr}   
\IEEEoverridecommandlockouts
\usepackage{cite}
\usepackage{amsmath,amssymb,amsfonts}
\usepackage{textcomp}
\usepackage{xcolor}

\usepackage{kotex}
\usepackage{algorithm}          
\usepackage[noend]{algpseudocode} 
\usepackage{multirow}
\usepackage{balance}
\usepackage{amsthm}
\usepackage{amsmath}
\usepackage{setspace}
\usepackage[tight,scriptsize]{subfigure}
\usepackage{ctable}
\usepackage{lmodern}
\usepackage[T1]{fontenc}
\usepackage[]{psfrag}
\graphicspath{{./fig/}}
\usepackage[switch]{lineno}
\usepackage{url}
\usepackage{threeparttable}

\newcommand{\eg}{{e.g.}}
\newcommand{\ie}{{i.e.}}

\newcommand{\LUTunit}{{\it LUT-unit}}

\newtheorem{definition}{Definition}

\newcommand{\reviewcolor}{black}

\def\BibTeX{{\rm B\kern-.05em{\sc i\kern-.025em b}\kern-.08em
    T\kern-.1667em\lower.7ex\hbox{E}\kern-.125emX}}
\begin{document}
\title{BiQGEMM: Matrix Multiplication with Lookup Table For Binary-Coding-based Quantized DNNs\\
\thanks{*Both authors contributed equally to this work.}
}

\author{
\IEEEauthorblockN{Yongkweon Jeon*, Baeseong Park*, Se Jung Kwon, Byeongwook Kim, Jeongin Yun, and Dongsoo Lee}
\IEEEauthorblockA{Samsung Research, Seoul, Republic of Korea \\
\{dragwon.jeon, bpbs.park, sejung0.kwon, byeonguk.kim, ji6373.yun, dongsoo3.lee\}@samsung.com}
}

\maketitle


\begin{abstract}

The number of parameters in deep neural networks (DNNs) is rapidly increasing to support complicated tasks and to improve model accuracy.
Correspondingly, the amount of computations and required memory footprint increase as well.
Quantization is an efficient method to address such concerns by compressing DNNs such that computations can be simplified while required storage footprint is significantly reduced.
Unfortunately, commercial CPUs and GPUs do not fully support quantization because only fixed data transfers (such as 32 bits) are allowed.
As a result, even if weights are quantized (by a non-uniform quantization scheme) into a few bits, CPUs and GPUs may not access multiple quantized weights without memory bandwidth waste.
Success of quantization in practice, hence, relies on an efficient computation engine design, especially for matrix multiplication that is a basic computation engine in most DNNs.
In this paper, we propose a novel matrix multiplication method, called BiQGEMM, dedicated to quantized DNNs.
BiQGEMM can access multiple quantized weights simultaneously in one instruction.
In addition, BiQGEMM pre-computes intermediate results that are highly redundant when quantization leads to limited available computation space.
Since pre-computed values are stored in lookup tables and reused, BiQGEMM achieves lower amount of overall computations.
Our extensive experimental results show that BiQGEMM presents higher performance than conventional schemes when DNNs are quantized.
\end{abstract}

\begin{IEEEkeywords}
Model Compression, Deep Learning, Machine Learning, AI Inference, Quantization, GEMM, GEMV
\end{IEEEkeywords}

\section{Introduction}

\begin{figure}[htbp]
\centerline{\includegraphics[width=1.0\linewidth]{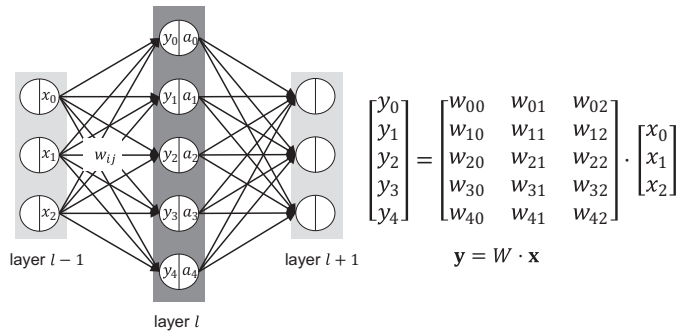}}
\caption{An example showing GEMV operations for layer $l$ (a fully connected layer). Given an input vector $\mathbf{x}\in\mathbb{R}^n$, which is the activation of the previous layer, and a weight matrix $W\in\{-1,1\}^{m\times n}$, an output vector $\mathbf{y}\in\mathbb{R}^m$ can be computed by using a GEMV routine, where $m$ and $n$ are output (hidden) size and input size, respectively. Activation $\mathbf{a}$ is the output of the activation function $f$ (\eg, sigmoid, tanh, and ReLU) with $\mathbf{y}$ as an input. In this manuscript, `output' refers to $\mathbf{y}$, not $\mathbf{a}$, unless specified otherwise.}
\label{f:intro}
\end{figure}

As the number of parameters in DNNs increases to improve model accuracy with various tasks, reducing inference latency is becoming more challenging.
Reducing response time becomes highly critical when real-time services are demanded (\eg, autonomous driving, automatic speech recognition, and neural machine translation). 
Note that most of response time is usually consumed by general matrix-to-matrix multiplication (GEMM) or general matrix-to-vector multiplication (GEMV) of high-order time complexity (see Fig.~\ref{f:intro}). 
Efficient computation of matrix multiplication, therefore, directly corresponds to response time reduction.
Previously in order to accelerate GEMM operations, both hardware- and software-based approaches have been introduced~\cite{li2019edge,reuther2019survey,choudhary2020comprehensive,cheng2017survey,zhou2019edge,he2018survey}. 

As an effort to reduce latency, few-batch multiplications\footnote{In this paper, we refer to either GEMV or GEMM as few-batch multiplication for convenience.} are strongly preferred for DNN inference at the cost of reduced weight reuse. 
Note that if GEMV is conducted to support single batch inference, weight matrix data is accessed only once.
Such a streaming-like operation is highly memory-bound in modern computing systems based on von Nuemann architecture, where main memory (DRAM) is separated from the processing unit~\cite{von1993first}.
Moreover, if weight matrix size becomes larger, then the portion of memory-bound operations is also larger.
Execution workloads with little data reuse on computing systems, therefore, would prevent an efficient utilization of computing resources because of the problem of memory-wall (also known as von Neumann bottleneck)~\cite{hennessy2011computer}. 
To alleviate a memory-access bottleneck from hardware perspectives, in-memory computing (in which computational operations are performed within the memory unit) has been widely studied~\cite{eleftheriou2018memory}. 
In other words, for DNNs, combating the memory bottleneck is desperate enough to request a new hardware architecture design paradigm.

As a practical solution at algorithm level, model compression is an effective technique to achieve lower end-to-end latency. 
Model compression reduces not only off-chip memory accesses (and hence, low power consumption) on mobile but also main memory bandwidth requirements by shrinking memory footprint with negligible accuracy drop~\cite{cheng2017survey,choudhary2020comprehensive}. 
Thus, model compression is being widely studied to accelerate inference computations. 
Popular model compression techniques include pruning~\cite{han2015deep,liu2018rethinking,lee2019network}, low-rank approximation~\cite{sainath2013low,li2016recovery}, and quantization~\cite{xu2018alternating,bhandare2019efficient,zhang2018lq}. 

In this work, we consider quantization because of its simple structure and high compression ratio~\cite{xu2018alternating,bhandare2019efficient,zhang2018lq}.
The rationale behind quantization for DNNs is that we can reduce the number of bits to represent each parameter without noticeable model accuracy \textcolor{\reviewcolor}{drop} because DNNs include a lot of redundancy.
Note that weights and activations need to be quantized at different times.
Weights are fixed during inference, and hence, weight quantization is performed in advance before performing inference.
On the other hand, activation quantization should be conducted on-the-fly with additional computations (for quantization) during inference.
If the quantization algorithm is complicated, then the cost of dynamic quantization might be larger than the gain from quantization effects.
In addition, activation compression may result in a serious accuracy degradation if training is not aware of a quantized structure of activations. 
In this manuscript, thus, we study weight quantization only that is enough to accelerate matrix multiplication as we demonstrate later. 

In this paper, we propose a novel matrix multiplication algorithm dedicated to quantized DNNs that can be performed on modern computer (von Neumann) architectures. 
Even though quantization obviously reduces storage requirements on off-chip memories, achieving higher performance with quantized DNNs on CPUs or GPUs is challenging.
\textcolor{\reviewcolor}{In particular, because data transfer on commercial processors is performed with a fixed width (such as 32 bits) while weights can be quantized with an arbitrary number of bits, accessing multiple (non-uniformly) quantized weights may cause some waste in bandwidth utilization. In order not to waste memory bandwidth, bit-packing is essential and unpacking process must be followed to multiply the sign (represented by each bit) and the scaling factor (each bit has a different scaling factor (See Fig.~\ref{fig:bg-comp-int8})), which is an overhead. In addition, decoding non-uniformly quantized weights may induce additional instructions. In other words, for binary-coding-based quantization, existing CPUs or GPUs may waste memory bandwidth or require additional special hardware.} Our proposed method, called BiQGEMM\footnote{non-\underline{GE}neral \underline{M}atrix to \underline{M}atrix multiplication for \underline{Bi}nary-coding-based \underline{Q}uantized neural networks}, addresses such concerns using lookup tables that accept quantized weights as indices.
BiQGEMM is built on the observation that quantization leads to a lot of redundant computations. 
The key idea is that for any real-number vector (of activations) $\mathbf{v}\in\mathbb{R}^{n}$, the number of possible outcomes from a dot product of $\mathbf{v}$ and a binary vector $\mathbf{b}\in\{-1,1\}^{n}$ (of quantized weights) is limited to be $2^n$, all of which can be pre-computed and stored in lookup tables that can be reused. 
We show that by replacing a majority of arithmetic operations with table lookups, BiQGEMM calculates matrix multiplications with high performance and improved bandwidth utilization.


\section{Background}\label{bg}
\subsection{Quantization}\label{bg:Qu}

\begin{figure}[t]
\centerline{\includegraphics[width=1.0\linewidth]{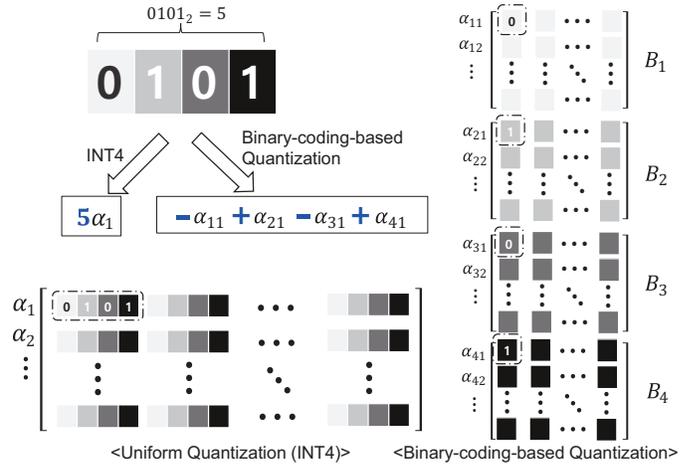}}
\caption{\textcolor{\reviewcolor}{A comparison on placement of each quantization bit between INT4 and Binary-coding-based quantization. Suppose that an element of a weight matrix is quantized by using 4 bits. Those 4 bits in a weight are continuously placed in INT4. On the other hand, for binary-coding-based quantization, quantized bits of one weight are distributed into 4 binary matrices. In INT4, each 4-bit vector represents a fixed-point number to be multiplied by a scaling factor while each bit in binary-coding quantization indicates the sign of a corresponding scaling factor. As such, when a weight is quantized as $0101_2$, how to dequantize such a binary number becomes vastly different depending on the selected quantization method.}}
\label{fig:bg-comp-int8}
\end{figure}

DNNs intentionally involve a lot of redundancy to expedite searching for an optimal local minimum.
Thus, the model size of DNNs has a potential to be significantly reduced by various compression algorithms.
Quantization is gaining increasing popularity as an effective model compression technique.
There exist various quantization formats and dequantized weights can be represented either by fixed-point numbers (based on uniform quantization) or by floating-point numbers (based on codebook lookups or binary-coding quantization).

Note that codebook-based quantization presents high compression ratios for various models with ignorable model accuracy degradation because expected values after quantization are still maintained to be floating-point numbers~\cite{facebook_lut_quant}.
Even though codebook-based quantization is highly efficient to reduce off-chip memory footprint, computational complexity is not reduced at all after dequantization.
Fixed-point quantization, on the other hand, reduces both storage requirement and computational complexity.
Since INT8 quantization associated with additional techniques is introduced to be able to maintain the model accuracy of some well-known CNN models, INT8 has been adopted by various commercial tools~\cite{bhandare2019efficient}.
Note that operations other than GEMV or GEMM need to be re-designed to function in INT8 while INT8-aware retraining may be necessary not to degrade model accuracy seriously.
For example, layer normalization and softmax operations for attention blocks for the Transformer demand floating-point computations~\cite{bhandare2019efficient}.
Accordingly, the overhead due to frequent conversions between fixed-point formats and floating-point formats might be close to the cost of quantized matrix multiplication~\cite{bhandare2019efficient}.


\begin{figure}[t]
\centerline{\includegraphics[width=0.95\linewidth]{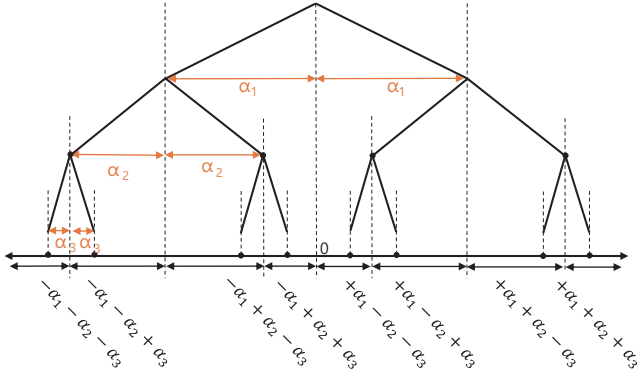}}
\caption{\textcolor{\reviewcolor}{An illustration of binary-coding-based quantization when the number of quantizaton bits is 3.}}
\label{fig:bcq}
\end{figure}

\begin{figure}[t]
\centerline{\includegraphics[width=0.95\linewidth]{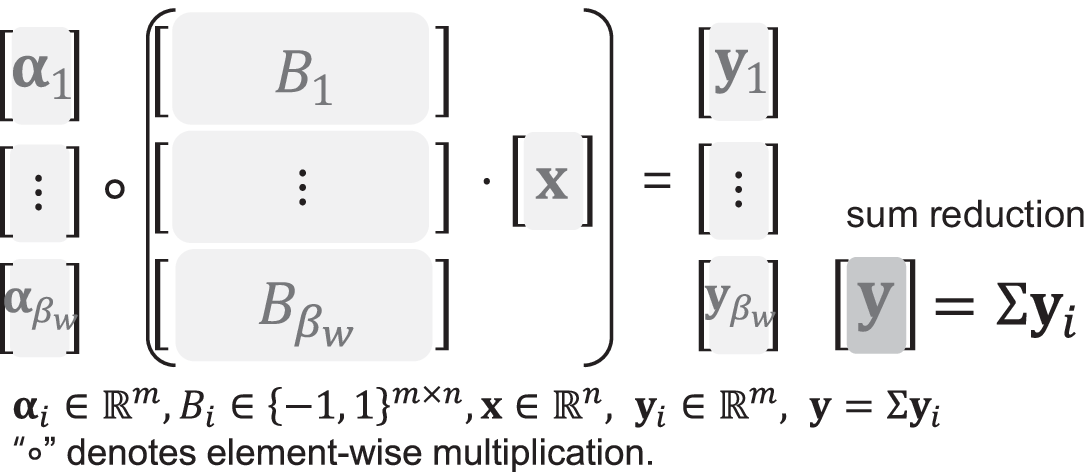}}
\caption{GEMV with multi-bit quantized weight matrix \textcolor{\reviewcolor}{(when quantized weights follow the structure of the binary codes).}}
\label{fig:bg-qbit}
\end{figure}

As an effort to reduce both computations and footprint significantly, binary-coding-based quantization has been proposed~\cite{xu2018alternating, rastegari2016xnor,zhang2018lq}.
Since expected values after binary-coding quantization remain to be floating-point numbers, accuracy degradation can be ignored even when only about 3 bits are used \textcolor{\reviewcolor}{for} quantization~\cite{xu2018alternating,zhang2018lq}.
Despite a possibility to highly simplify computations, binary-coding-based quantization has not been useful in practice because a computing system should allow bit-level memory access.
In this manuscript, hence, we consider binary-coding-based quantization as a baseline to enable practical usages in commercialized computing systems.

Note that in the case of INT8, activations should be also quantized in order to allow fixed-point GEMV or GEMM, while such activation quantization with floating-point-based quantization is optional.
Activation quantization inherently demands dynamic quantization process during inference.
Even though inference operations can be a lot more efficient by previously proposed methods such as method of 4 Russian~\cite{aho1974design} or popcount- and XOR-logic~\cite{rastegari2016xnor}, activation quantization requires 1) modifications to training algorithm to severely restrict the range of activation values and 2) computational overhead for format conversions~\cite{rastegari2016xnor,bhandare2019efficient,xu2018alternating}.
In this paper, we show that BiQGEMM presents high performance even when activations are maintained to be floating-point numbers.

\subsection{Binary-coding-based quantization}


When a real-number vector $\mathbf{w}\in\mathbb{R}^{p}$ is quantized into $q$ bits by binary-coding-based quantization method, $\mathbf{w}$ is mapped into scaling factors $\alpha_i$ $\in\mathbb{R}$ and binary vectors $\mathbf{b}_{i}\in\{-1,+1\}^{p}$ ($1\leq i \leq q$).
Then, $\mathbf{w}$ is approximated as $\sum_{i=1}^{q} \alpha_i \mathbf{b}_{i}$ where scaling factors are shared by multiple elements in $\mathbf{w}$.
Scaling factors and binary vectors are obtained as follows:
\begin{align}
    \arg\min_{\alpha_i,\mathbf{b}_i}||\mathbf{w}-\sum_{i=1}^{q} \alpha_i\mathbf{b}_i||^2
    \label{eq:mse}
\end{align}
such that quantization error is minimized.
Since there is no analytical solution to minimize such quantization error, numerous heuristic approaches have been proposed~\cite{guo2017network,xu2018alternating,zhang2018lq,courbariaux2016binarized,amc}.

The same principle of binary-coding-based vector quantization can be applied to matrices where quantization can be independently performed for each row or column.
For a weight matrix quantized into binary matrices $B_i$ with scaling factor vectors $\boldsymbol{\alpha}_i$, multiplication with a real-number vector $\mathbf{x}$ produces an output vector $\mathbf{y}$ as follows:

\begin{align}
\mathbf{y}=\sum_{i=1}^{\beta_w} (\boldsymbol{\alpha}_i \circ (B_i\cdot\mathbf{x})) \label{eq:weight}
\end{align}

\noindent where operation $\circ$ denotes element-wise multiplication (\ie, Hadamard product) and $\beta_w$ is the number of quatization bits for weights.
Fig.~\ref{fig:bg-qbit} illustrates how to perform multiplication of multi-bit quantized weight matrices by a real-number vector. 
Note that for convenience, binary weight matrices $B_i$s can be concatenated in vertical direction and multiplied by a vector $\mathbf{x}$. 
Then, element-wise multiplication by scaling factor $\mathbf{\alpha}_i$ produces an intermediate partial output $\mathbf{y}_i$.
Finally, sum of vectors of $\mathbf{y}_i$ yields the final output $\mathbf{y}$. 

Consider that a real-number activation vector $\mathbf{x}$ is also quantized by using $\beta_a$ bits into $\mathbf{s}_j\in\{-1,1\}^{n}$ with scaling factors $\gamma_j$ ($1\leq j \leq \beta_a$), the previous output $\mathbf{y}$ now can be computed as follows:

\begin{align}
\mathbf{y}=\sum_{i=1}^{\beta_w} (\boldsymbol{\alpha}_i \circ (B_i\cdot \sum_{j=1}^{\beta_a} \gamma_j\mathbf{s}_j)).
\label{eq:both}
\end{align}

\noindent Eq.~\ref{eq:both} suggests that activation quantization would increase the number of computations compared to Eq.~\ref{eq:weight}, even though most computations are simple as bit-wise logic. 
It should be noted that without sophisticated hardware design support for bit-wise logic incurred by binary-coding-quantization, activation quantization may degrade matrix multiplication performance.

\subsection{Natural Language Processing}\label{bg:nlp}

In order to set a range of parameters (such as matrix size) to be used for our experiments and to take into account the impact of the proposed algorithm, we investigate natural language processing (NLP) as a representative application of BiQGEMM. 

RNNs~\cite{rumelhart1986learning} and the Transformer~\cite{vaswani2017attention} are being widely accepted as time-series data analysis tools to process natural language tasks. 
Long short-term memory (LSTM)~\cite{hochreiter1997long}, compared with conventional RNNs, introduces additional gates in a unit to overcome long-term dependency and gradient vanishing problem in vanilla RNNs.
As such, most recently proposed RNN-based networks employ LSTM as a basic unit to improve model accuracy on multiple benchmark language models. 
The Transformer presented another noticeable advance in NLP. 
By breaking the recurrent structure and fully exploiting attention mechanism~\cite{bahdanau2014neural}, the Transformer better figure out the relevance between words in the sentences.
Correspondingly, the Transformer has been initially proposed for neural machine translation (NMT), and then extended to a wide range of NLP tasks, including BERT~\cite{devlin2018bert}, with impressive results on GLUE~\cite{wang2018glue} and SQUAD~\cite{rajpurkar2016squad}.

The structure of the Transformer can be divided into encoder layers and decoder layers.
An encoder layer includes one attention block structured as four ($n\times n$) weight matrices and a feed-forward block with ($n\times 4n$) and ($4n\times n$) matrices, where $n$ is the hidden size. 
Also, a decoder layer presents two attention blocks and a feed-forward block while the structure of each block is the same as that of the encoder. 
The number of encoder layers is chosen to be 6 (6) and $n$ is selected to be 512 (1024) for the base (big) model.
Weight matrices are fed into matrix multiplication operations and the weight matrix size is rapidly increasing to support various complicated tasks with increased model accuracy goals.
For example, for NMT, most models that show excellent performance are based on the big model version of the Transformer~\cite{ahmed2017weighted,shaw2018self,ott2018scaling,edunov2018understanding}. 
T5, another variant of the Transformer, increases the number of weights to 11 billion and the number of layers to 24~\cite{Raffel2019ExploringTL}. 

BERT~\cite{devlin2018bert} is a pre-training-based model for applications that require only the encoder part of the Transformer. 
BERT models are known to continuously set new records on model accuracy with high number of encoder layers and hidden size (such as 24 and 1024, respectively). 
Associated with new training algorithms based on the large model of BERT, various advanced models, such as XLNet~\cite{yang2019xlnet}, RoBERTa~\cite{liu2019roberta}, and ERNIE~\cite{zhang2019ernie,sun2019ernie}, are being developed. 
Ever increasing requests of higher accuracy demands only larger weight matrices.
For instance, the biggest weight matrix size in xx-large model of ALBERT~\cite{lan2019albert} is ($4K\times 16K$), which requires 256 MB (with FP32) of memory footprint. 
Such large weight matrices cannot avoid frequent DRAM accesses even if the same parameters are repeatedly reused over the whole network.

As for automatic speech recognition (ASR), similarly, the number of parameters is also increasing to accomplish higher model accuracy~\cite{karita2019comparative,karita2019improving,irie2019choice,luscher2019rwth,park2019specaugment,han2019state}. 
To illustrate, LAS is an end-to-end ASR DNN model based on bi-directional LSTM using six encoder layers with ($2.5K\times 5K$) weight matrix structure and two decoder layers with ($1.2K\times 1.2K$) weight matrix structure~\cite{park2019specaugment}. 

In sum, fast matrix multiplication with a matrix size of (at least) a few thousands is essential to realize DNNs of NLP tasks.
Such high-performance matrix multiplication needs to assume that DNNs are compressed because of increasing number of parameters.

\subsection{Quantizing the Transformer}\label{bg:tr}

\begin{table}
\renewcommand{\arraystretch}{1.3}

\centering
\caption{Quantization Comparison on the Transformer}
\scriptsize
\begin{threeparttable}
\begin{tabular}{clcl}
\hline
\multirow{2}{1cm}{} & \multirow{2}{1cm}{Models} & Data format (bits) & English-to-German\tabularnewline
 &  & (weight / activation) & BLEU\tabularnewline
\cline{1-4}
\multirow{2}{1cm}{Ref~\cite{bhandare2019efficient}} & Baseline & 32 / 32 & 27.68\tabularnewline
\cline{2-4}
 & Uniform & 8 / 8 & 27.30 (-0.22)\tabularnewline
\cline{1-4}
\multirow{4}{1cm}{Ref~\cite{prato2019fully}} & Baseline & 32 / 32 & 26.46\tabularnewline
\cline{2-4}
 & \multirow{3}{1cm}{Uniform} & 8 / 8 & 26.38 \textcolor{\reviewcolor}{(-0.08)}\tabularnewline
 &  & 6 / 6 & 26.98 (+0.52)\tabularnewline
 &  & 4 / 4 & 18.32 (-8.14)\tabularnewline
\cline{1-4}
\multirow{5}{1cm}{Ref~\cite{deeptwist}} & Baseline & 32 / 32 & 25.8\tabularnewline
\cline{2-4}
 & \multirow{4}{1cm}{Binary-Coding\\(Greedy)} & 4 / 32 & 25.5 (-0.3)\tabularnewline
 &  & 3 / 32 & 25.3 (-0.5)\tabularnewline
 &  & 2 / 32 & 23.9 (-1.9)\tabularnewline
 &  & 1 / 32 & 0.4 (-25.4)\tabularnewline
\cline{1-4}
\end{tabular}
\end{threeparttable}
\label{tab:transofrmer}
\end{table}

\begin{table}
\renewcommand{\arraystretch}{1.3}
\centering
\caption{Memory Usage with Different Number of Quantization Bits (Weights: $512$-by-$512$ matrix, batch size: 18)}
\scriptsize
\begin{threeparttable}
\begin{tabular}{|c|c|c|c|c|c|c|}
\hline
\multicolumn{3}{|c|}{Data format (bits)} & \multicolumn{4}{c|}{Memory (MB)}\tabularnewline
\cline{1-7}
W & A & O & W & A & O & total\tabularnewline
\hline
32 & 32 & 32 & 1.049 & 0.037 & 0.037 & 1.122\tabularnewline
\hline
8 & 8 & 32 & 0.262 & 0.009 & 0.037 & 0.308\tabularnewline
\hline
6 & 6 & 32 & 0.197 & 0.007 & 0.037 & 0.240\tabularnewline
\hline
4 & 4 & 32 & 0.131 & 0.005 & 0.037 & 0.173\tabularnewline
\hline
\hline
4 & 32 & 32 & 0.131 & 0.037 & 0.037 & 0.205\tabularnewline
\hline
3 & 32 & 32 & 0.098 & 0.037 & 0.037 & 0.172\tabularnewline
\hline
2 & 32 & 32 & 0.066 & 0.037 & 0.037 & 0.139\tabularnewline
\hline
\end{tabular}
\begin{tablenotes}
\item[] W: weights, A: activations (inputs), O: outputs.
\end{tablenotes}
\end{threeparttable}
\label{tab:mem}%
\end{table}%

Now we estimate the number of quantization bits using the Transformer that are being widely applied to various NLP tasks.
Table~\ref{tab:transofrmer} lists quantization results of the (base model) Transformer (designed to perform English to German translation) using uniform quantization~\cite{bhandare2019efficient,prato2019fully} and binary-coding-based quantization with greedy approximation~\cite{guo2017network}. 
For uniform quantization results, we refer to the numbers from~\cite{bhandare2019efficient,prato2019fully}.
For binary-coding-based quantization based on greedy approximation method (to reduce quantization error), we retrain the model using quantization-aware training algorithm introduced in~\cite{deeptwist} using WMT13 data set.
When retraining the model, all hyper-parameters are the same as in the Transformer~\cite{vaswani2017attention} except large initial learning rate by 2$\times$ and additional hyper-parameter of distortion step (introduced in~\cite{deeptwist}) that is set to be 2000.
The baselines results for each quantization case are inherently different due to different initialization conditions and test set, and the number of quantization bits and translation quality (given as BLEU score) are described in Table~\ref{tab:transofrmer}. 

Table~\ref{tab:mem} shows the memory usage when weights and activations are quantized into different number of bits while a matrix size is fixed to be 512-by-512 (that is the size of an attention layer of the base Transformer). 
The number of sub-words in the test data set is 18 on average, and thus, batch size is 18.
Note that because of relatively small dimension of activations, activation quantization does not reduce memory footprint as much as weight quantization, while more bits for weight quantization need to be assigned given a target model accuracy as shown in Table~\ref{tab:transofrmer}.
Such observation is consistent with other matrix sizes. 
Combining Table~\ref{tab:transofrmer} and Table~\ref{tab:mem}, for BiQGEMM design considerations, we quantize only weights while we are mainly interested in a few bits for quantization (\eg, 1 to 3 quantization bits).

\section{Methodology}\label{ME}
\subsection{Motivation and Definitions}


\begin{definition}
\LUTunit~$\mu$ is the length of a sub-vector to be used as an input argument of a table lookup function.
\end{definition}

\begin{definition}
Given an $m$-by-$n$ matrix denoted by $A$, $A^{r}_{\mu}$ is a $\mu$-by-$\frac{m\times n}{\mu}$ matrix reshaped from $A$ while maintaining column-wise traversal.
\end{definition}

\begin{definition}
Given an $m$-by-$n$ matrix denoted by $A$, $A[i;j]$ is a sub-matrix of $A$ formed by $i$-to-$j$ columns when $i\leq j<n$.
\end{definition}

\begin{definition}
Given a column vector $\mathbf{v}$ of length $n$, $\mathbf{v}[i;j]$ is a sub-vector of $\mathbf{v}$ comprised of $i$-to-$j$ rows when $i\leq j<n$.
\end{definition}

\begin{definition}
$M_\mu\in\{-1, 1\}^{2^\mu\times\mu}$ denotes a matrix constructed by concatenating all possible (non-redundant) $2^{\mu}$ binary vectors of $\mu$ length.
\end{definition}

\begin{figure}[t]
\centerline{\includegraphics[width=0.95\linewidth]{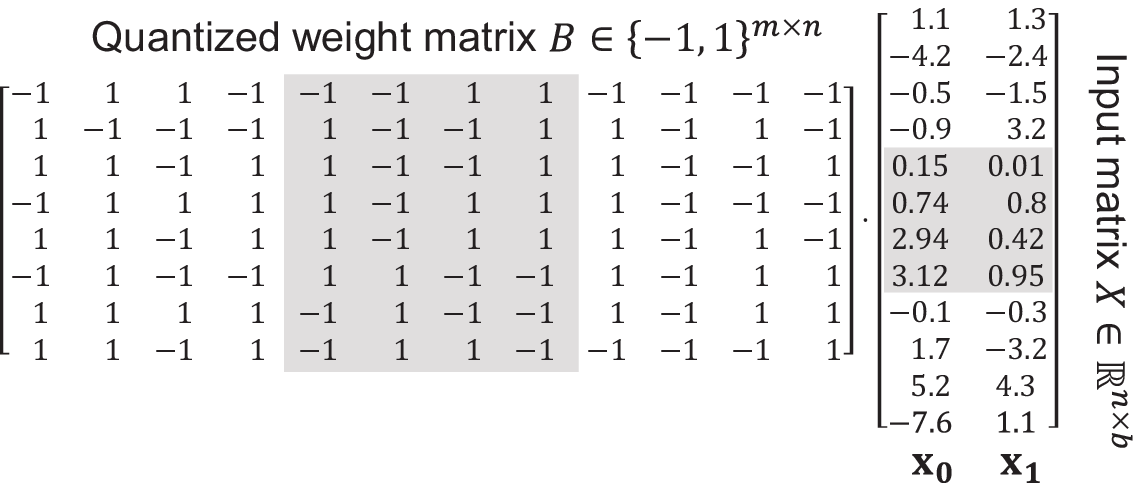}}
\caption{An example of quantized weight matrix $B$ and input matrix $X$ composed of two input vectors $\mathbf{x_0}$ and $\mathbf{x_1}$.}
\label{fig1}
\end{figure}

\noindent We assume that a binary weight matrix $B\in\{-1,1\}^{m\times n}$ and an input matrix $X\in\mathbb{R}^{n\times b}$ are given, where $m$, $n$, and $b$ are output size, input size, and batch size, respectively.
Fig.~\ref{fig1} shows an example of a quantized (binary) weight matrix and an input matrix. In Fig.~\ref{fig1}, each matrix is equally divided into three parts along with \textit{LUT-unit} $\mu$ of 4. 
Considering a shaded part in Fig.~\ref{fig1}, a row vector (having 4 binary digits) in $B[4;7]$ is one of $2^\mu$ possible combinations. 
Correspondingly, each row after a product of $B[4;7]$ and $\mathbf{x_0}[4;7]$ is also limited to be one of $2^\mu$ possible vectors. 
As an attempt to exploit a strictly limited space of available outputs, the product of $M_\mu$ and reshaped input matrix $X_{\mu}^{r}$ is pre-computed and stored in lookup tables. 
Then, pre-computed values are retrieved from lookup tables using $\mu$-bit binary digits in a weight matrix as a key (or an index). 
Note that when output size is large enough (i.e., $2^{\mu}\ll m$), computing efficiency is enhanced because most arithmetic operations can be replaced by retrieval operations. 



\begin{figure}[t] 
	\centering
	\subfigure[Construction of lookup tables. Each column vector $\mathbf{q}_i^j$ in $Q$ represents a lookup table corresponding to $\mathbf{x}_i^j$.] {
		\includegraphics[width=0.95\linewidth]{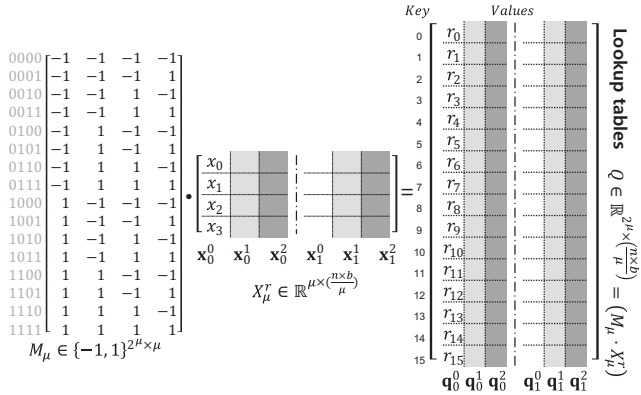}
		\label{fig:lut-a}
	}
	\subfigure[Dynamic programming method to build a lookup table.] {
		\includegraphics[width=0.95\linewidth]{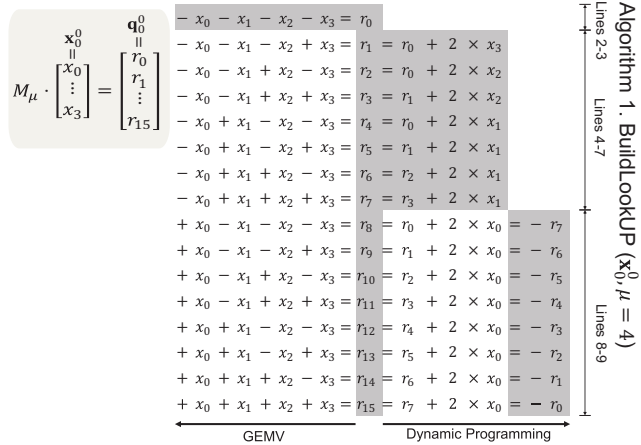}
		\label{fig:lut-b}
	}
	\caption{Illustration of two different methods to build lookup tables when \LUTunit~$\mu$ is 4.}
	\label{fig:lut}
\end{figure}

\subsection{Algorithm Description}


When \LUTunit~$\mu$ is given as a parameter, each column in the product of $M_{\mu}$ and $X^{r}_{\mu}$ becomes entries of a separate lookup table. 
Fig.~\ref{fig:lut} shows an exemplary process of building lookup tables when $\mu$ is $4$, where we define a sub-vector $\mathbf{x}_\alpha^\beta$ of length $\mu$ and a lookup table $\mathbf{q}_\alpha^\beta$ corresponding to $\mathbf{x}_\alpha^\beta$ as follows:

\begin{align}
\mathbf{x}_\alpha^\beta\overset{\Delta}{=}\mathbf{x}_\alpha[\mu\beta; \mu\beta+\mu-1]
\end{align}
\begin{align}
\mathbf{q}_\alpha^\beta\overset{\Delta}{=}M_\mu\cdot\mathbf{x}_\alpha^\beta
\end{align}

\noindent when $0\leq\alpha<b$ and $0\leq\beta < \frac{n}{\mu}$, where $b$, $n$, and $\mu$ are the batch size, the input size, and the \LUTunit, respectively.
Then, the product of a sub-matrix $B[\mu\beta;\mu\beta+\mu+1]$ and a sub-vector $x_\alpha^\beta$ can be found in lookup table $\textbf{q}_\alpha^\beta$, instead of performing GEMV. In other words, partial products of GEMM are replaced with table lookups in BiQGEMM.


\begin{algorithm}[htbp]
\footnotesize
\caption{\textsc{Build a lookup table with dynamic programming}}
\label{a-build}
\renewcommand\algorithmicrequire{\textbf{Input}:}
\renewcommand\algorithmicensure{\textbf{Output}:}
\begin{algorithmic}[1]
\Require \LUTunit~$\mu\in\mathbb{N}$ ($\mu\ll m$), where $m$ is output size
\Require a sub-vector $\mathbf{x}_\alpha^\beta=\{x_0, x_1,...,x_{\mu\texttt{-}1}\}\in\mathbb{R}^{\mu}$,
\Ensure A lookup table ($\mathbf{q}_\alpha^\beta$) 
\vspace{0.15cm}
\Procedure{BuildLookUP}{$\mathbf{x}_\alpha^\beta$, $\mu$}
\For {$i\gets 0$ to $\mu\texttt{-}1$}
    \State $r_0 \gets r_0+x_i$
\EndFor
\State $k \gets 1$
\For {$i\gets 1$ to $\mu\texttt{-}1$}
	\For {$j\gets 0$ to $2^{i-1}\texttt{-}1$}
    	\State $r_{k\texttt{++}} \gets r_j + 2\times x_{i\texttt{-}1}$ 
    \EndFor
\EndFor
\For {$i\gets 1$ to $2^{\mu\texttt{-}1}$}
	\State $r_{2^\mu\texttt{-}i} \gets -r_{i\texttt{-}1}$
\EndFor
\EndProcedure
\end{algorithmic}
\end{algorithm}

As shown in Fig.~\ref{fig:lut-b}, building a lookup table can be optimized by using dynamic programming technique.
Specifically, while constructing a lookup table $\mathbf{q}_0^0$, dynamic programming reduces redundant arithmetic operations (described as right-sided equations in Fig.~\ref{fig:lut-b}) compared to the case when GEMV using $M_{\mu=4}$ and $\mathbf{x}_0^0$ is performed (described as left-sided equations in Fig.~\ref{fig:lut-b}). 
Algorithm~\ref{a-build} presents the pseudo code of building a lookup table with dynamic programming. 
In Fig.~\ref{fig:lut-b}, each equation is annotated with the corresponding line numbers in Algorithm~\ref{a-build}. 
Note that every $n/\mu$ sub-vector per input induces a distinct lookup table of $2^\mu$ entries, and hence, the time complexity of the proposed dynamic programming scheme $T_{c,dp}$ is calculated as follows:

\begin{align}
T_{c,dp}=O((2^\mu+\mu-1)\cdot \frac{n}{\mu}\cdot b)\approx O(2^\mu\cdot\frac{n}{\mu}\cdot b).
\label{eq.tcdp}
\end{align}

\noindent $T_{c,dp}$ obtained by our proposed technique is $\mu$ times less than $T_{c,mm}=O(2^\mu\cdot\mu\cdot\frac{n\times b}{\mu})$ that is time complexity of GEMM-based lookup table construction method (see Fig.~\ref{fig:lut-a}).
Suppose our dynamic programming scheme is combined with multi-threading, each thread is responsible for constructing one or more lookup tables (\ie, one lookup table cannot be implemented by coordinating more than two threads).
Another level of parallelism is achieved by vectorizing independent equations in Fig.~\ref{fig:lut-b} to utilize SIMD instructions.
Note that because of dependency among equations in the case of dynamic programming, however, conventional GEMV or GEMM schemes might be a better choice to fill up lookup table entries if a computing system embeds a sizable number of simple computation units (\eg, GPU). 
In other words, depending on the characteristics of a processor to run BiQGEMM, a choice of appropriate scheme to implement lookup tables would be different.



\begin{figure}[t]
\centerline{\includegraphics[width=0.95\linewidth]{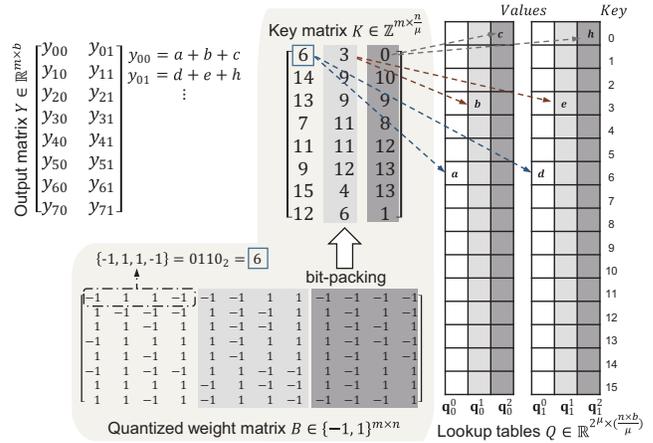}}
\caption{Illustrations of retrieving partial results from lookup tables.}
\label{fig:lut-c}
\end{figure}

Fig.~\ref{fig:lut-c} shows an illustrated process of querying partial results from lookup tables. 
Consecutive \LUTunit~$\mu$ binary data in a quantized weight matrix $B$ are grouped into a sub-vector that can be converted into an integer number $k$ where $0\leq k < 2^{\mu}$ (\eg, $\{ -1, 1, 1, -1\}$ is converted into an integer 6 when $\mu$ is 4). 
In other words, the key matrix $K$ in Fig.~\ref{fig:lut-c} is a $\mu$-bit-packed version of $B$, and each element in $K$ serves as an index of table lookups\footnote{Note that matrix $K$ instead of $B$ can be loaded in advance into the system, since the weight matrices are fixed during inference.}. 
Partial results retrieved from lookup table entries are accumulated for each sub-vector of length $\mu$ per input vector, and the BiQGEMM is completed. 

\begin{figure}[t]
\centerline{\includegraphics[width=.75\linewidth]{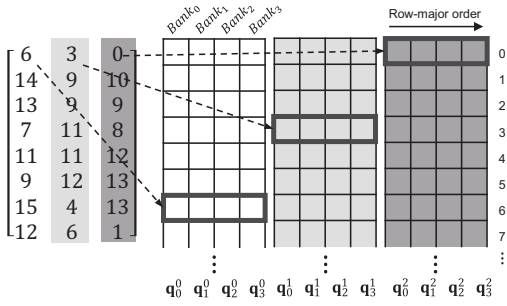}}
\caption{Lookup tables arrangement considering SIMD operations when the batch size $b$ is 4.} 
\label{fig:simd}
\end{figure}

All keys in the key matrix are used $b$ (=the number of input vectors) times.
For example, all of the leftmost lookup tables corresponding to every input (\ie, $q_0^0$ and $q_1^0$ in Fig.~\ref{fig:lut-c}) are commonly accessed by the first column of key matrix. 
In other words, different lookup tables are accessed by a shared key.
By continuously placing lookup table entries associated with the same key as shown in Fig.~\ref{fig:simd}, SIMD operations in CPU are encouraged, and bank conflicts in GPU can be mitigated. 
As for GPU, scratchpad memory or software controlled caches (\ie, shared memory in CUDA) can store lookup tables. 
Then, even if the memory accesses are irregular, multiple threads can fetch multiple data in parallel unless multiple addresses within the same memory bank are accessed. 
Thus, a penalty of irregular access to a lookup table on GPU is not as critical as that of CPU.

\begin{figure}[t]
\centerline{\includegraphics[width=0.95\linewidth]{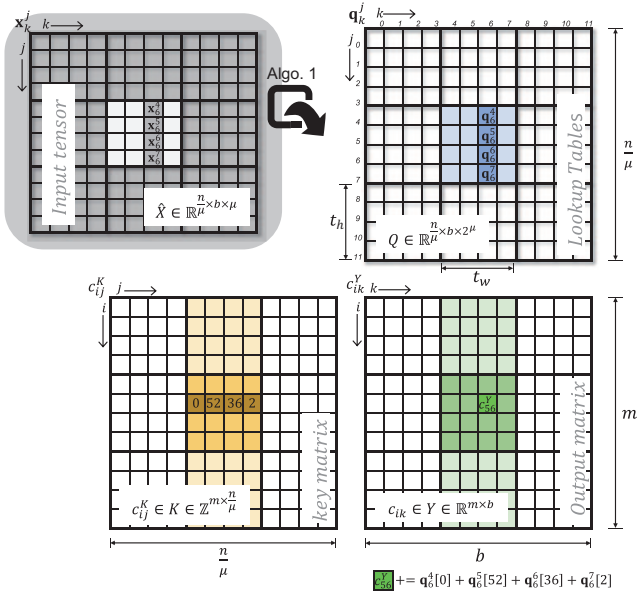}}
\caption{A tiling strategy for LUT-stationary BiQGEMM. Each cell in $K$ and $Y$ represents a scalar, while each cell in $\hat{X}$ and $Q$ implies a sub-vector $\mathbf{x}_{i}^{j}$ and a lookup table $\mathbf{q}_{i}^{j}$, respectively, where $\hat{X}$ is a reshaped one from an input matrix $X$ (\ie, 3-dimensional representation of $X_\mu^r$). $t_w$ and $t_h$ are tile width and height, respectively.}
\label{fig:tiled}
\end{figure}

\begin{algorithm}[htbp]
\footnotesize
\caption{\textsc{LUT-based inference with LUT-stationary tiling}} 
\label{a-query}
\renewcommand\algorithmicrequire{\textbf{input}:}
\renewcommand\algorithmicensure{\textbf{output}:}
\begin{algorithmic}[1]
\Require Key Matrix ($K\in\mathbb{Z}^{m\times\frac{n}{\mu}}$), Input Tensor ($\hat{X}\in\mathbb{R}^{\frac{n}{\mu}\times b\times\mu})$
\Ensure Output Matrix($Y\in\mathbb{R}^{m\times b}$) \Comment{See Figure~\ref{fig:tiled}}
\Procedure{QueryLUT}{$\hat{X}$, $K$}
\For{each tile $T_X$ in $\hat{X}$}
    \State build Lookup tables $T_Q\subset Q$ {\it w.r.t.} $T_X$ \Comment{Algorithm~\ref{a-build}}
    \For{each pair of Tiles $(T_K\subset K, T_Q\subset Q)$ }
        \For{each cell $c_{ij}^K\in T_K$ } 
            \State $c_{ik}^{Y} \gets c_{ik}^{Y} + \mathbf{q}_k^j[c_{ij}^{K}]$
        \EndFor
    \EndFor
\EndFor
\EndProcedure
\end{algorithmic}
\end{algorithm}

A tiling approach for BiQGEMM is illustrated in Fig.~\ref{fig:tiled} and described in Algorithm~\ref{a-query}.
To build a lookup table (LUT) without redundancy, BiQGEMM adopts an LUT-stationary tiling scheme. 
Two input arguments of BiQGEMM are given as a key matrix $K$ and an input tensor $\hat{X}$ reshaped from an input matrix (while the reshaped form is determined by a user-defined parameter $\mu$). 
For tiling, tile width $t_w$ and height $t_h$ need to be specified. 
Each tile in LUT is operated by one thread (that is assigned to one SM in the case of GPU), and hence, multiple accesses to lookup tables in a block can be SIMDified (by multiple CUDA cores in the case of GPU). 
Lookup tables in $T_Q\subset Q$ are not constructed in advance (\ie, not taken from DRAM), instead, implemented on-the-fly by Algorithm~\ref{a-build} with sub-vectors $\mathbf{x}_k^j$ as inputs (\textit{Line 3} in Algorithm~\ref{a-query}). 
After implementing lookup tables in $T_Q$, pairs of tiles in $Q$ and $K$ corresponding to $T_Q$ are loaded successively and individually, and computed by using $T_Q$ (\textit{Lines 4--6} in Algorithm~\ref{a-query}). 
With LUT-stationary tiling, partial output results obtained by multiple threads are processed through either sum reduction or atomic additions to obtain the final output values. 
Since each of $m\cdot\frac{n}{\mu}$ keys is utilized $b$ times, the worst-case time complexity required to retrieve $T_r$ is given as follows:

\begin{align}
T_r=O(m\cdot\frac{n}{\mu}\cdot b).
\label{eq.trn}
\end{align}

To process multi-bit quantization of weight matrices, BiQGEMM assumes that multiple binary weight matrices are concatenated as described in Fig.~\ref{fig:bg-qbit}. 
Note that such concatenation does not increase the number of lookup tables, and thus for BiQGEMM, only an amount of lookup table retrieving operations increases as the number of quantization bits increases.
In short, for multi-bit quantized weight matrices, $T_r$ becomes $O(m\cdot\frac{n}{\mu}\cdot b\cdot\beta)$, where $\beta$ is the number of quantization bits.



\subsection{Complexity Analysis}\label{me:ca}

Given an input matrix $X\in\mathbb{R}^{n\times b}$ and a quantized binary weight matrix $B\in\{-1,1\}^{m\times n}$, a matrix multiplication $Y=B\cdot X$ performed by GEMM yields $O(m\cdot n\cdot b)$ as time complexity, where $m$, $n$, and $b$ are output size, input size, and batch size, respectively (Fig.~\ref{f:intro} shows an example when $b$ is 1).
Time complexity analysis on a matrix multiplication performed by BiQGEMM, on the other hand, is divided into the following two parts: i) constructing lookup tables (Eq.~\ref{eq.tcdp}) and ii) retrieving lookup table entries (Eq.~\ref{eq.trn}).
Correspondingly, time complexity of BiQGEMM $T$ is presented as follows:
\begin{align}
T=T_{c,dp}+T_r & =O(2^\mu\cdot\frac{n}{\mu}\cdot b+m\cdot\frac{n}{\mu}\cdot b)
\\& =O(m\cdot n\cdot b\cdot (\frac{2^\mu+m}{m\cdot\mu}))\label{etn}
\end{align}

\noindent If $2^\mu\ll m$ is satisfied, then $T$ can be approximated as

\begin{align}
T\approx O(\frac{m\cdot n\cdot b}{\mu}).
\label{tn}
\end{align}

\noindent Note that as long as $2^\mu\ll m$, Eq.~\ref{tn} holds regardless of a choice of algorithm to build lookup tables (\ie, irrespective of a selection between $T_{c,dp}$ or $T_{c,mm}$). 
Then, by using BiQGEMM instead of conventional GEMM, time complexity of a matrix multiplication is reduced by $\mu$. 
For multi-bit quantized weights, time complexity of both BiQGEMM and GEMM increases linearly with the number of quantization bits.


Since the underlying principles of BiQGEMM are fundamentally different compared to GEMM, rethinking hardware designs is necessary.
First, while performance of FMA units is directly related to GEMM performance, the usage of FMAs for BiQGEMM is limited to constructing lookup tables.
Second, while cache design is useful for GEMM to utilize spatial locality in SRAM when loading a matrix by accessing successive data, BiQGEMM cannot efficiently facilitate such a locality because accessing entries of lookup tables would be non-sequential in general (note that, nonetheless, such degraded locality is not fatal if BiQGEMM is associated with multi-batch inference on CPU or with scratchpad on GPU (see Fig.~\ref{fig:simd})). 
In addition, because BiQGEMM is desired to produce lookup tables (that are usually larger than an input matrix) to be placed in SRAM, an available range of tile size would be highly constrained compared to GEMM. 
Now, let us explain why BiQGEMM is designed to be efficiently operated in CPUs or GPUs with quantized weights despite such issues (\ie, low utilization of FMAs and low data access locality).
Note that with quantized weights, GEMM needs to decompress bit-packed quantized weight data by 1) performing two-step operations to extract quantized weights bitwise from a much wider data container (such as INT32) and 2) conducting two-step arithmetic operations to convert data of the form of $\{0,1\}$ into the form of $\{-1,1\}$ (see Algorithm~\ref{a-unpack} and Fig.~\ref{fig:bg-comp-int8}).
On the other hand, BiQGEMM directly accesses and utilizes the bit-packed weight data as keys (or indices) of lookup tables without such additional decompressing steps.
It should be noted that for quantized weights, overhead by decompression can outweigh the gain by the reduced memory footprint as we demonstrate in the next section.

\begin{algorithm}[htbp]
\footnotesize
\caption{\textsc{Unpacking for GEMM}} 
\label{a-unpack}
\renewcommand\algorithmicrequire{\textbf{input}:}
\renewcommand\algorithmicensure{\textbf{output}:}
\begin{algorithmic}[1]
\Require packed data $x$, 
\Ensure unpacked weight $\mathbf{w}\in\{-1,1\}^{32}$
\Procedure{unpacking}{$x$}
\For {$i\gets 0$ to $31$} 
    \State $w_i \gets (((($x$ >> i) \& (1))*2)-1$
\EndFor
\EndProcedure
\end{algorithmic}
\end{algorithm}

Consideration of existing hardware architecture designs is one of the keys to understanding the impacts of BiQGEMM on the system performance.
For example, even though large tile size for BiQGEMM would result in improved data reuse, current CPU or GPU designs allow limited tile size of BiQGEMM such that large batch size (\ie, compute-bound workload) might be less favorable to BiQGEMM. 
New hardware design dedicated to BiQGEMM is, therefore, suggested as an important future work.

\section{Experimental Results}\label{ER}


\subsection{Setup}
We implemented our proposed algorithm BiQGEMM in C++/CUDA with various compilers targeting different processor architectures. 
Table~\ref{tab:Machines} presents descriptions of systems where tests are performed. 
As for tests with CPUs, performance of BiQGEMM is compared with Intel MKL (\textsc{mkl})~\cite{wang2014intel}, Eigen (\textsc{eigen})~\cite{eigenweb}, FBGEMM (\textsc{int8f})\cite{FBGEMM,FBGEMMD}, QNNPACK (\textsc{int8q})~\cite{QNNPACK,QNNPACKD} and an algorithm introduced in~\cite{weissteinmatrix} (\textsc{kCpu}). 
Additionally BiQGEMM is also run by GPU and compared with cuBLAS (\textsc{cublas})~\cite{nvidia2008cublas}, a kernel code in CUDA samples (\textsc{kGpu})~\cite{volkov2008benchmarking}, and XNOR-popcout (\textsc{xnor})~\cite{courbariaux2016binarized}. Note that we included \textsc{int8f}, \textsc{int8q}, and \textsc{xnor} for the purpose of comparison on performance even though model accuracy and required hardware designs are all different. 
We generated synthetic matrices filled by random numbers as data sets for tests.

\begin{table}
\renewcommand{\arraystretch}{1.3}
\centering
\caption{Machine configurations used in Section~\ref{ER}}
\scriptsize
\begin{threeparttable}
\begin{tabular}{|l|ccc|}
\hline
 & Mobile & PC & GPGPU\tabularnewline
\hline
Processor & Cortex-A76 & i7-7700 & Tesla v100\tabularnewline
\# of Cores & 4 & 4 & 80 (SMs)\tabularnewline
L1D-cache & 64KB/core & 32 KB/core & 128 KB/SM\tabularnewline
SIMD lane & 4/core & 8/core & 16*4/SM\tabularnewline
DRAM & 8GB & 16 GB & 16 GB\tabularnewline
GB/s\tmark[*] & 31.8 & 35.76 & 900 \tabularnewline
FLOPS & 19.36G $\times$ 4 & 57.6G $\times$ 4 & 181.87G $\times$ 4\tabularnewline
\hline
Compiler & LLVM 8.0.7 & gcc 5.4.0 & nvcc 10.2\tabularnewline
OS & Android 9 (4.14.78) & \multicolumn{2}{c|}{ubuntu 16.04.6 (4.15.0)}\tabularnewline
\cline{1-4}
\end{tabular}
\begin{tablenotes}
\item[*] Maximum memory bandwidth
\end{tablenotes}
\end{threeparttable}
\label{tab:Machines}%
\end{table}%


Our proposed algorithm accepts \LUTunit~$\mu$ as a user-defined parameter that can affect system performance.
Let us explain how \LUTunit~$\mu$ is optimized in practice.
$\mu$ determines physical space allocated for lookup tables. 
If $\mu$ increases, then the number of lookup tables decreases while the number of entries in each lookup table increases exponentially (see Fig.~\ref{fig:lut-a} and Eq.~\ref{eq.tcdp}) (\ie, there exists a trade-off between the number of LUTs and the number of entries inside each LUT). 
Combined with output size $m$, \LUTunit~$\mu$ specifies a relative performance gain of BiQGEMM over GEMM.
Specifically from Eq.~\ref{etn}, for a given output size $m$, we can find $\mu$ by $\mathrm{arg\,min}_{\mu}\frac{2^\mu+m}{m\cdot\mu}$.
Note that in practical situations, hardware resources may limit the maximum $\mu$ (due to internal SRAM size), and thus, restrict tile size as well.
Thus, theoretically optimized $\mu$ should be verified empirically throughout extensive experiments.
We use $\mu=8$ (for our entire tests) that turns out to be close to the value optimized in theory .

\subsection{BiQGEMM Runtime Profiling}

\begin{figure}[t] 
	\centering
	\subfigure[$n=1K$, $b=32$] {
		\includegraphics[width=0.46\linewidth]{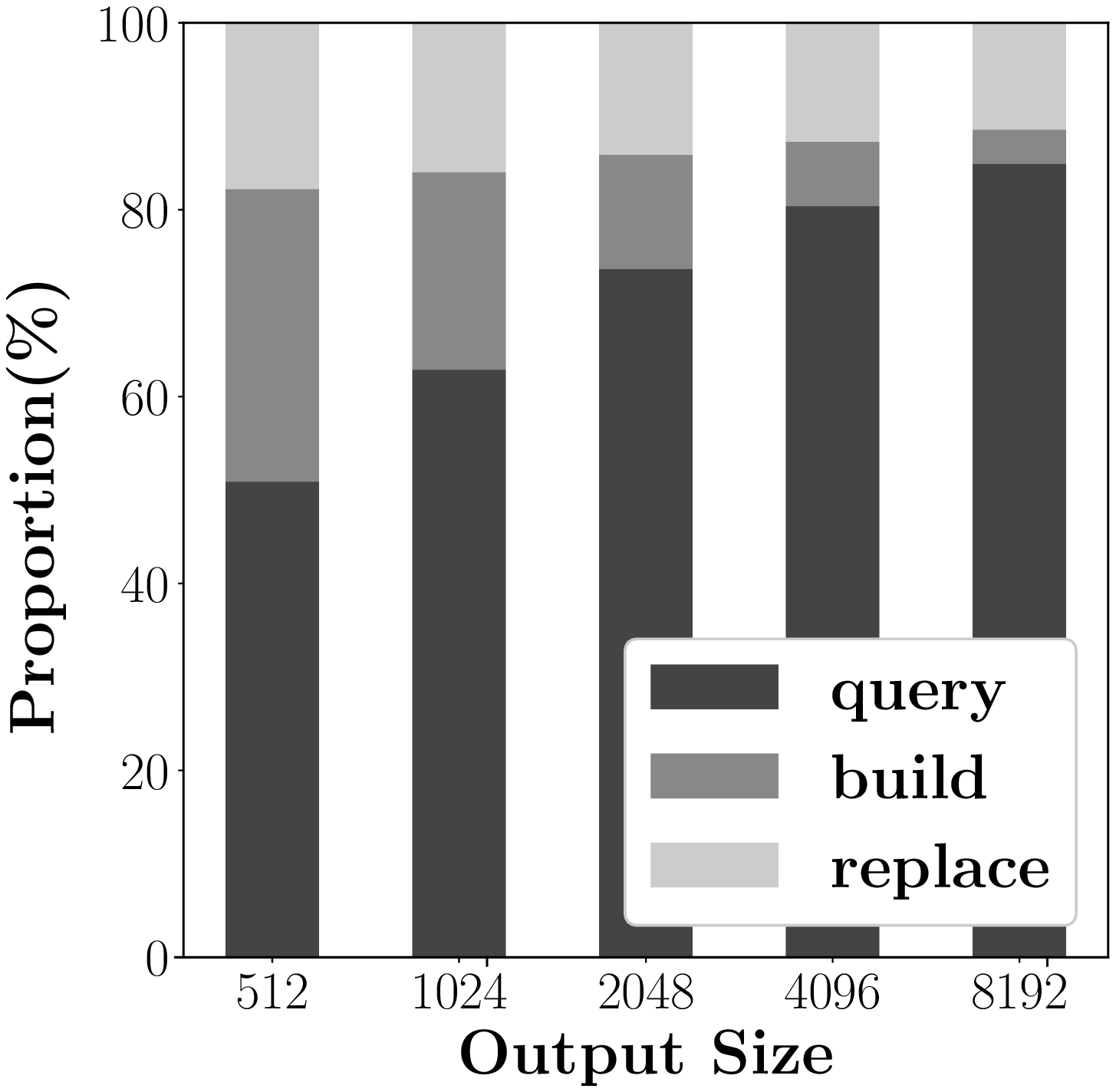}
		\label{ex:prof-a}
	}
	\subfigure[$n=2K$, $b=32$] { 
		\includegraphics[width=0.46\linewidth]{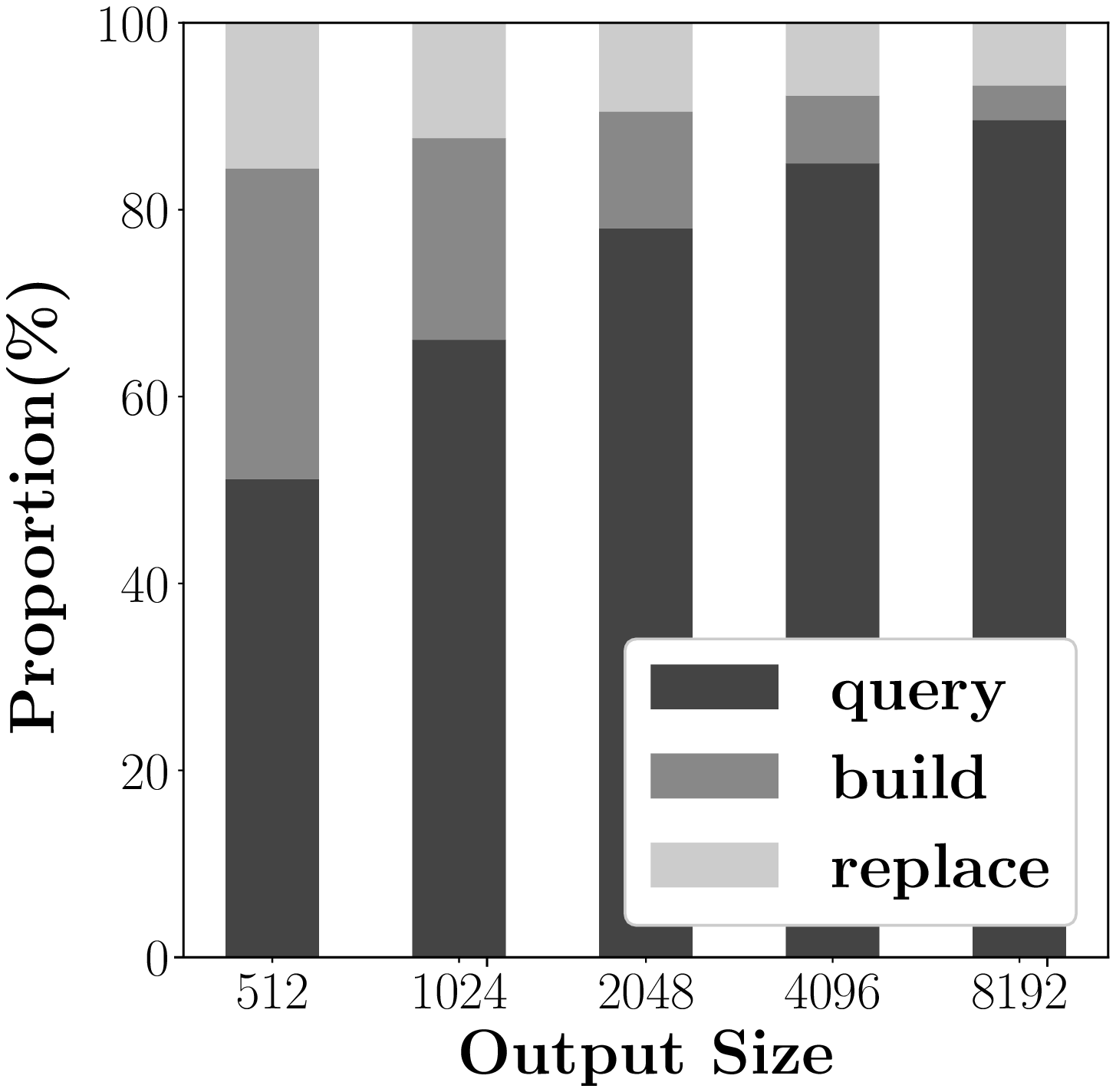}
		\label{ex:prof-b}
	}
	\caption{Runtime profiling of BiQGEMM. As output size $m$ increases, the portion of retrieving (query) operations also increases. For all matrix sizes selected, retrieving operations are dominating the entire performance.}
	\label{ex:portion}
\end{figure}

Fig.~\ref{ex:portion} represents the runtime portion of each operation when running BiQGEMM on a CPU with various output size $m$.
Operations are mainly categorized into 1) lookup tables construction (\textit{build}), 2) retrieving values (\textit{query}), and 3) memory replacement for tiling (\textit{replace}).
As discussed in Section~\ref{me:ca}, increasing output size induces a larger proportion in the process of retrieving values, and correspondingly, more arithmetic operations in GEMM to be replaced with retrieval operations in BiQGEMM. 
Note that even when more quantization bits are assigned to each weight, BiQGEMM increases the retrieving operations only which are relatively inexpensive among 3 operations (see Fig.~\ref{fig:bg-qbit}). 
As such, when weights are quantized, BiQGEMM is better implemented (for performance compared with a GEMM-based scheme) when output size is larger and weights are quantized with more bits.

\subsection{GEMM with Quantized and Bit-packed Weights}\label{ss:packing}

\begin{figure}[t] 
	\centering
	\subfigure[on CPU] {
		\includegraphics[width=0.46\linewidth]{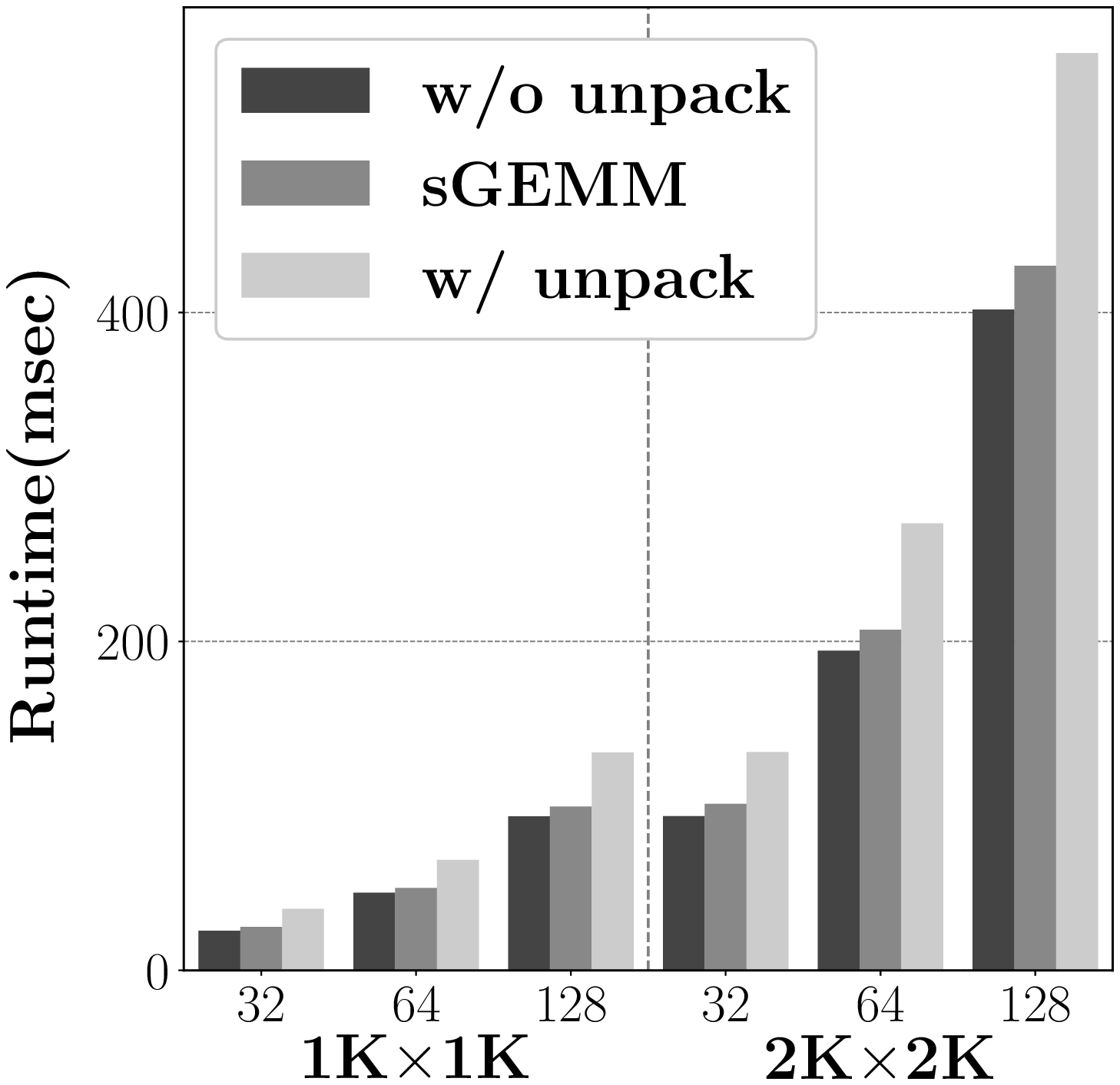}
		\label{ex:pack-a}
	}
	\subfigure[on GPU] {
		\includegraphics[width=0.46\linewidth]{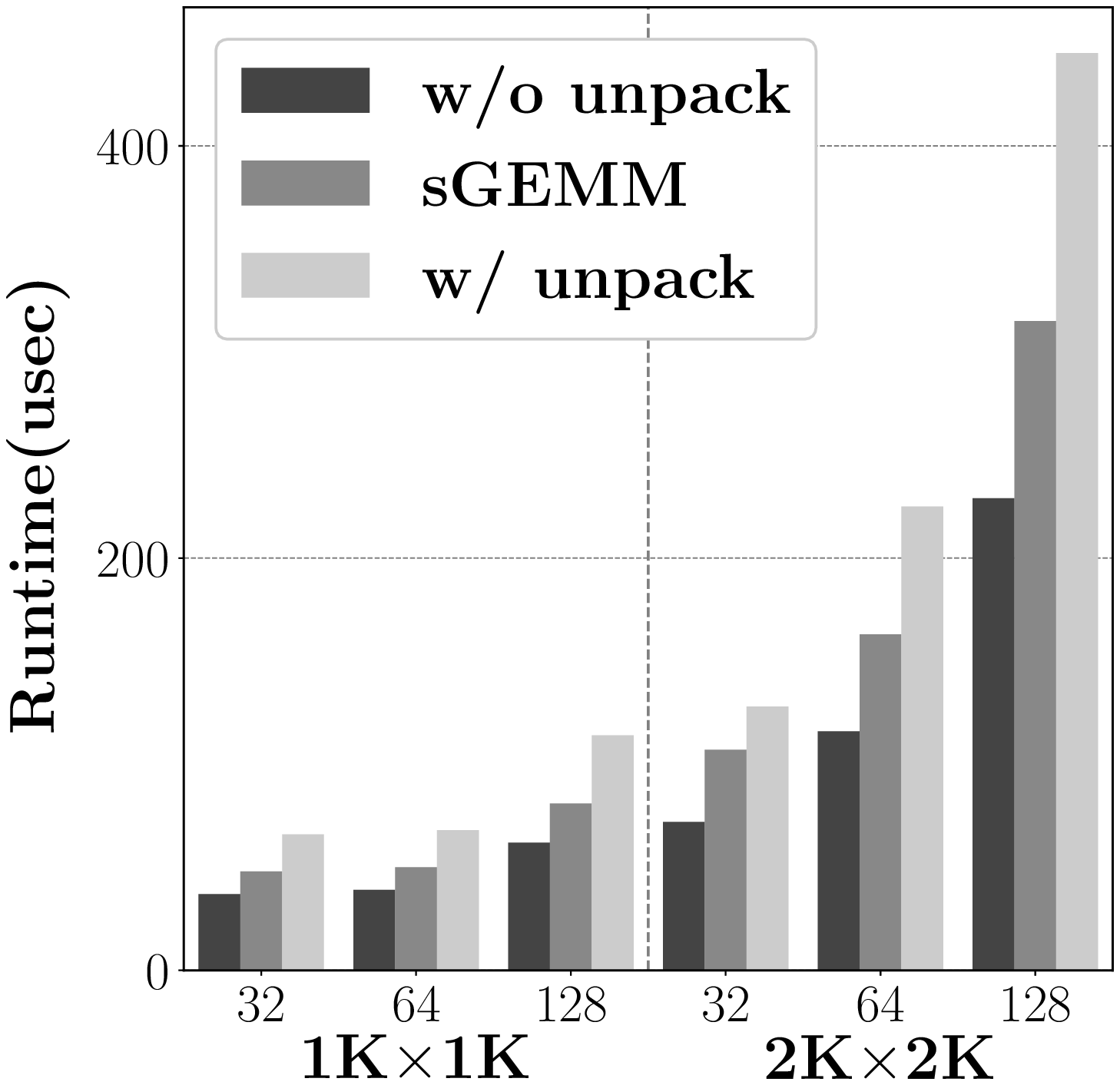}
		\label{ex:pack-b}
	}
	\caption{The plots representing overhead incurred by unpacking bits when weights are 1-bit quantized, and \textsc{kCpu}~\cite{weissteinmatrix} and \textsc{kGpu}~\cite{volkov2008benchmarking} are used. The weight matrices used in these experiments are square matrices of order $m (=n)$ with batch size 32, 64, and 128. To ensure fair comparison, the same compiler optimization is applied on the codes.}
	\label{ex:pack}
\end{figure}

To reduce memory footprint, bit-packing is essential for quantized models to be densely stored in a general data type, such as INT32.
Through bit-packing with few batch multiplication, memory-bound algorithms are accelerated by reduced memory bandwidth requirements.
However, unpacking is required to be performed prior to running GEMM operations with packed data.
Since unpacking fundamentally requires bit-level manipulations, unpacking on CPUs and GPUs may cause a large computational overhead.
Indeed, Fig.~\ref{ex:pack} demonstrates such concerns.
Assuming that weights are 1-bit quantized, Fig.~\ref{ex:pack} compares runtime of 3 different scenarios (w/o unpack, sGEMM, and w/ unpack) depending on how the binary vectors are processed: `sGEMM' indicates a case where only one bit is stored in a 32-bit container while `w/ unpack' means multiplying bit-packed data after extracting bits through unpacking process.
Since `sGEMM' does not assume bit-packed data formats, quantization would not affect performance (\ie, performance would be the same as that of full-precision weights).
Note that `w/o unpack' measures runtime when bit-packed data is multiplied by a real vector without unpacking (\ie, products of a 32-bit packed data (a scalar) and a vector of length 32), which will produce incorrect result, but is useful to identify performance gain by decreased memory access latency.
Runtime gap between `w/o unpack' and `sGEMM' implies performance gain by reduced memory footprint, whereas the difference between `w/o unpack' and `w/ pack' runtime indicates latency overhead by unpacking operations.
Fig.~\ref{ex:pack} confirms that GEMM with quantized weight is inefficient in terms of the response time even though quantization reduces DRAM memory access latency.  

\subsection{Comparison with others}
\begin{figure*}[t] 
	\centering
	\subfigure[PC (i7-7700)] {
        \includegraphics[width=1.0\linewidth]{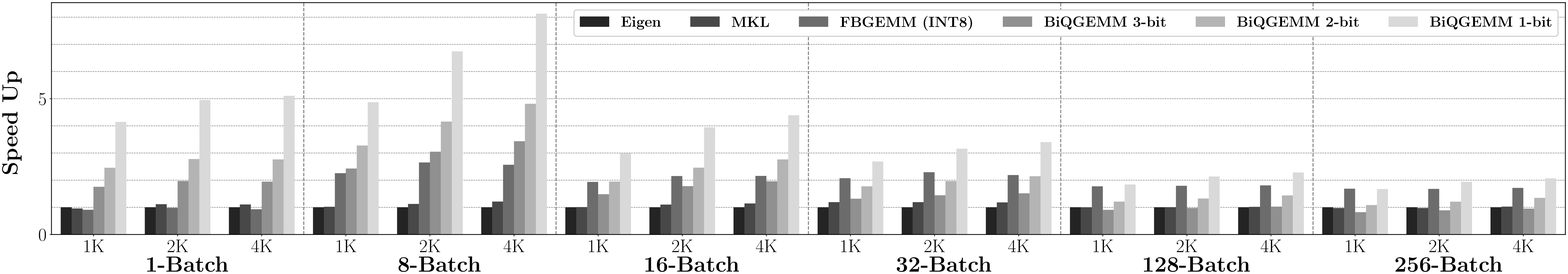}
		\label{ex:i7}
	}
	\subfigure[\textcolor{\reviewcolor}{Mobile (Cortex-A76)}] {
		\includegraphics[width=1.0\linewidth]{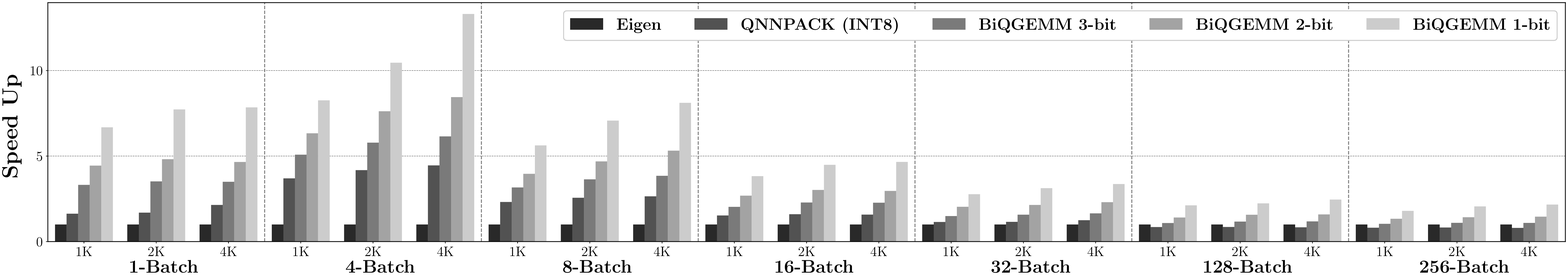}
		\label{ex:s10}
	}
	\caption{Speedup over \textsc{eigen} using 1-thread. Matrix size is given as $m$-by-$1K$. Output size $m$ and batch size are annotated along the horizontal axis.}
	\label{ex:speedup}
\end{figure*}


Even though small batch size is preferred for inference to reduce response time, recently developed DNNs demand batch size to be larger than 1.
For example, an input (in the form of a sequence) of Transformers' encoder contains several sub-words (tokens) to detect hidden relationships between sub-words. 
Because all of the sub-words in the input are multiplied by the same weights concurrently, those sub-words are processed in a group manner. 
The number of sub-words, thus, is equivalent to batch size in terms of computation.
Accordingly, we conduct experiments using various batch sizes ranging from 1 to 256 considering the number of sub-words used for the Transformer and its variants.

Since unpacking process adds significant overhead to run GEMM-based schemes with quantized bit-packed weights as shown in Section~\ref{ss:packing}, `sGEMM' version (that stores only one bit in a 32-bit containers without packing) introduced in the previous subsection is selected to be compared with BiQGEMM.
`sGEMM' version does not benefit from quantization, and therefore, 1-bit quantized weights and full-precision weights result in the same performance measured when using MKL (\textsc{mkl}) and Eigen (\textsc{eigen}) (thus, we do not specify whether weights are quantized in Fiq.~\ref{ex:speedup} for \textsc{eigen} and \textsc{mkl}). 
Performance of BiQGEMM is measured when weights are quantized into 1, 2, or 3 bits.
Note that the runtime of both BiQGEMM and GEMM with quantized weights linearly increases as the number of quantization bits increases. 
Then, combined with an observation that \textit{BiQGEMM 1-bit} (BiQGEMM with 1-bit quantization) shows highest performance in Fig.~\ref{ex:speedup}, BiQGEMM is always faster than GEMM given the same quantization bits. 
Moreover, if batch size is small enough, then BiQGEMM performance with 2- or 3-bit quantization outperforms not only GEMM with full-precision but also GEMM with INT8\footnote{Unlike weight matrix, quantization and packing process of activations (inputs) need to be performed during inference time, and thus, included in the elapsed times of \textsc{int8f} and \textsc{int8p}.}. 
Thus, even if the latency is of the top priority in the inference system design (at the expense of increased memory footprint without quantization), BiQGEMM can be the selection when the number of quantization bits is small enough with allowable model accuracy degradation.


Fig.~\ref{ex:speedup} shows that when input size is fixed, larger output size enhances speedup of BiQGEMM significantly because of higher reuse rate of lookup tables and correspondingly increased number of arithmetic operations to be replaced by simple table lookups.
Large batch size, on the other hand, may have adverse effects on BiQGEMM performed by CPUs or GPUs.
Fig.~\ref{ex:speedup} describes that BiQGEMM can be slower than the other GEMM kernels if batch size and the number of quantization bits are beyond a certain threshold value.
In theory, since time complexity of BiQGEMM is given as $O(\frac{\beta\cdot m \cdot n \cdot b}{\mu})$ and $\mu$ is empirically optimized to be 8, BiQGEMM with under 8-bit quantization is supposed to be always faster than GEMM (of full-precision) regardless of batch size. 
However, in reality, we need to consider limiting factors due to available hardware resources as discussed in Section~\ref{ME}. 
If batch size increases and data reuse is improved correspondingly, then overall computational efficiency improvement of \textsc{mkl}, \textsc{eigen}, and \textsc{int8f} can be higher than that of BiQGEMM.
For example, when batch size exceeds 128 in Fig.~\ref{ex:i7}, \textsc{eigen} and \textsc{mkl} are faster than BiQGEMM with 3-bit quantization.
Specific batch size determining whether BiQGEMM can be faster than GEMM depends on the system configuration.
For example, in the case of mobile CPU with low computational power (see Table~\ref{tab:Machines}), BiQGEMM outperforms not only full-precision GEMM (\textsc{eigen}) but also INT8 GEMM (\textsc{int8q}) even when batch size becomes larger compared to the case of PC as described in Fig.~\ref{ex:s10}.

Even though experimental results in Fig.~\ref{ex:speedup} assumed only one thread, multi-threading linearly improves performance of both BiQGEMM and GEMM that can be parallelized by tiling techniques.




\begin{table}[t]
\renewcommand{\arraystretch}{1.3}
\centering
\caption{Runtime comparison on GPGPU}
\scriptsize
\begin{threeparttable}
\begin{tabular}{|c|c|c|c|c|c|}
\hline
weights & \multirow{2}{1cm}{batch\\size} & \multicolumn{4}{c|}{runtime ($\mu$sec)}\tabularnewline
\cline{3-6}
(n-by-n) &  & BiQGEMM & \textsc{kGpu} & \textsc{cublas} & \textsc{xnor}*\tabularnewline
\cline{1-6}
\multirow{4}{1cm}{512} & 1 & 4 & 22 & 12 & 18\tabularnewline
\cline{2-6}
 & 32 & 11 & 24 & 20 & 18\tabularnewline
\cline{2-6}
 & 128 & 30 & 39 & 25 & 19\tabularnewline
\cline{2-6}
 & 256 & 58 & 63 & 26 & 19\tabularnewline
\cline{1-6}
\multirow{4}{1cm}{1024} & 1 & 4 & 36 & 14 & 18\tabularnewline
\cline{2-6}
 & 32 & 20 & 57 & 27 & 19\tabularnewline
\cline{2-6}
 & 128 & 70 & 120 & 45 & 21\tabularnewline
\cline{2-6}
 & 256 & 135 & 204 & 64 & 24\tabularnewline
\cline{1-6}
\multirow{4}{1cm}{2048} & 1 & 5 & 93 & 31 & 19\tabularnewline
\cline{2-6}
 & 32 & 47 & 153 & 52 & 23\tabularnewline
\cline{2-6}
 & 128 & 175 & 366 & 109 & 29\tabularnewline
\cline{2-6}
 & 256 & 330 & 661 & 179 & 40\tabularnewline
\cline{1-6}
\multirow{4}{1cm}{4096} & 1 & 7 & 213 & 90 & 23\tabularnewline
\cline{2-6}
 & 32 & 130 & 614 & 130 & 34\tabularnewline
\cline{2-6}
 & 128 & 528 & 1396 & 339 & 64\tabularnewline
\cline{2-6}
 & 256 & 1005 & 2516 & 594 & 109\tabularnewline
\cline{1-6}
\end{tabular}
\begin{tablenotes}
\item[*] It includes the packing cost for inputs (activations), but not the quantization cost for input. 
\end{tablenotes}
\end{threeparttable}
\label{tab:gpu}%
\end{table}%

\subsection{Experiments with GPU}

On GPU, we compare BiQGEMM with ~\textsc{kGpu},~\textsc{cuBLAS}, and \textsc{xnor}. 
Both \textsc{cublas} and~\textsc{kGpu} assume that only 1 bit is occupied in 32-bit containers (with unnecessary storage of 31 bits) for each quantized weight (\ie, bit-packing is not considered because unpacking is as slow as sGEMM). 
In the case of \textsc{xnor}, activations are quantized as well such that matrix multiplications are mainly computed by XNOR and popcount operations without unpacking procedure.
Assuming weights and activations are $\beta_{w}$- and $\beta_{a}$-bit quantized, \textsc{xnor} shows time complexity of $O(\beta_{w}\cdot\beta_{a}\cdot (m\cdot\frac{n}{32}\cdot b))$, where $m$, $n$, and $b$ are output size, input size, and batch size, respectively.
Although activation quantization can simplify computations further, activation quantization demands training algorithm modifications and computational overhead during inference as discussed in Section~\ref{bg} at the cost of model accuracy drop.

\textsc{cublas} is provided as a library form developed by a chip vendor such that we select ~\textsc{kGpu} as a baseline that we modify to implement BiQGEMM (for \textsc{xnor}, we use a publicly available code that is also based on \textsc{kGpu}).
Table~\ref{tab:gpu} shows runtime with various matrix sizes and batch sizes when each weight is 1-bit quantized (for \textsc{xnor}, activations are also 1-bit quantized).
Difference in performance between BiQGEMM and \textsc{kGpu} represents the gain by reduced bandwidth requirement and by improved computational principles. 
As shown in Table~\ref{tab:gpu}, BiQGEMM is faster than~\textsc{kGpu} by 1.08$\sim$30.42 times (as weight matrix size increases and batch size decreases, BiQGEMM becomes relatively faster).
Note that if batch size is small enough (to be memory-bound), even compared with \textsc{xnor}, BiQGEMM presents the best performance.



\section{\textcolor{\reviewcolor}{Discussion}}\label{DC}

\textcolor{\reviewcolor}{
Let $C~(m\times n)$ be the product of two matrices $A~(m\times k)$ and $B~(k\times n)$. 
BiQGEMM presents high performance in accelerating such a matrix multiplication operation especially when $m$ is large and $n$ is small (when $m$ and $n$ correspond to an output size and a batch size, respectively, in this manuscript). Note that for well known DNNs performing NLP tasks (including Transformers and LSTM), most layers can be represented as matrices (to be computed by GEMM) along with large $m$ and small $n$.
In convolutional neural networks (CNN), operations for convolutions can be transformed into GEMM routines as follows: When a batch size, an output feature map size, a filter size, a number of input channels, and a number of output channels are given as $b$, $h_o \times w_o$, $h_f \times w_f$, $c_i$, and $c_o$, respectively, then $m$, $k$, and $n$ correspond to ($c_o$), ($c_i \times h_f \times w_f$), and ($b\times h_o\times w_o$), respectively~\cite{chetlur2014cudnn}.
Correspondingly, compared to NLP tasks, CNNs usually yield relatively small $m$ and relatively large $n$.
As such, BiQGEMM would not be the best choice for CNNs if latency is the major concern. 
In addition, activations are also required to be quantized in CNNs to reduce memory footprint because the amount of activations are usually larger than that of weights. For CNNs, thus, it would be necessary to consider INT8 operation or XNOR-popcnt accompanied by compressing both weights and activations.}

\textcolor{\reviewcolor}{
There are a variety of issues (with different priority) on inference implementation that we need to consider to decide the optimal compression technique.
For instance, lowering end-to-end latency may be of the utmost importance while reducing memory footprint can be critical. 
Also, an acceptable degree of accuracy drop highly affects the choice of a particular model compression method.
Some networks are very sensitive to activation quantization, but others may not.
As such, a large spectrum of model compression techniques is demanded to support optimizing various aspects of inference implementation.
In this regard, BiQGEMM is able to enlarge such a spectrum. 
For DNNs associated with NLP tasks, BiQGEMM based on the binary-coding-based quantization can be a reasonable solution to accelerate computations while shrinking memory footprint even without the need to compress activations.}

\section{Conclusion}\label{CC}

We proposed an efficient matrix-to-matrix multiplication technique dedicated to quantized neural networks.
When weights are quantized, available output space of computational results is quite limited such that for a large matrix multiplication, a lot of computations become redundant.
BiQGEMM removes such redundancy by replacing multiplications with table lookups. 
Moreover, because commercial processors enable only a fixed data transfer width, a lot of memory bandwidth may be wasted when weights are non-uniformly quantized into a few bits.
BiQGEMM provides a way to access multiple quantized weights simultaneously regardless of the number of quantization bits.
Hence, memory bandwidth utilization is enhanced by BiQGEMM significantly while required memory bandwidth is reduced by quantization.
We demonstrated that BiQGEMM is a lot faster than previous matrix multiplication schemes especially when matrix size is large and batch size is small.


\bibliographystyle{ieeetr}
\bibliography{manuscript}

\end{document}